\documentclass{svproc}
\usepackage{url}

\usepackage[numbers]{natbib}
\usepackage{multicol}
\usepackage[bookmarks=true]{hyperref}
\usepackage{graphicx}
\usepackage{amsmath}
\usepackage{float}
\usepackage[dvipsnames]{xcolor}
\usepackage{colortbl}

\usepackage{caption}
\usepackage{subcaption}
\usepackage{siunitx}
\usepackage{booktabs}

\definecolor{Gray}{gray}{0.85}
\definecolor{LightCyan}{rgb}{0.88,1,1}
\definecolor{antiquewhite}{rgb}{0.98, 0.92, 0.84}

\newcolumntype{a}{>{\columncolor{Gray}}c}
\newcolumntype{d}{>{\columncolor{GreenYellow}}c}
\newcolumntype{p}{>{\columncolor{antiquewhite}}c}

\DeclareMathOperator*{\minimize}{Minimize}
\DeclareMathOperator*{\subjectto}{Subject\ to}
\DeclareMathOperator*{\FK}{FK}
\newcommand{\norm}[1]{\left\lVert#1\right\rVert}

\makeatletter

\newcommand{\Rmnum}[1]{\expandafter\@slowromancap\romannumeral #1@}
\makeatother

\begin{document}
\mainmatter              
\title{Time Variable Minimum Torque Trajectory Optimization for Autonomous Excavator}
%

%
\author{Yajue Yang$^{1}$, Jia Pan$^{2}$, Pinxin Long$^{3}$, Xibin Song$^{3}$, Liangjun Zhang$^{3}$}

\titlerunning{Trajectory Optimization for Autonomous Excavator}

\authorrunning{Yajue Yang et al.} 
%
%
\institute{City University of Hong Kong
\and University of Hong Kong \and Robotics and Auto-Driving Lab, Baidu Research
}

\maketitle              

\begin{abstract}
In this paper, we present a minimal torque and time variable trajectory optimization method for autonomous excavator considering the soil-tool interaction. The method formulates the excavation motion generation as a trajectory optimization problem and takes into account geometric, kinematic and dynamics constraints. To generate time-efficient trajectory and improve the overall optimization efficiency, we propose a time variable trajectory optimization mechanism so that the time intervals between the keypoints along the trajectory subject to the optimization. As a result, the method uses few keypoints and reduces the total number of optimization variables. We further introduce a soil-tool interaction force model, which considers the geometric shape of the bucket and the physical properties of the soil. The experimental result on a high fidelity dynamic simulator shows our method can generate feasible trajectories, which satisfy excavation task constraints and are adaptive to different soil conditions.

\end{abstract}

\section{Introduction}
Heavy machinery such as excavators are widely used in construction, mining and many other scenarios. There is a strong need for automating the excavation process in order to save the labor cost and improve the overall working condition, especially for hazardous environment. The fundamental function of an excavator is to remove or reshape materials by excavation operations. During a digging cycle, interaction forces between materials and the bucket might require large torque of the excavator joints to complete an excavation task. Thus, for preventing mechanical damage and saving fuel consumption, it is necessary to generate a trajectory with small torque demands. In addition, high productivity requires the trajectory to be time-efficient. Even though excavation motions appear highly repetitive, automating the excavator to meet these performance criteria still remains as a challenging problem. This motivates us to develop a motion planner that generates optimal excavation trajectories simultaneously considering dynamics and time factors.

\begin{figure}
    \centering
    \includegraphics[width=\linewidth]{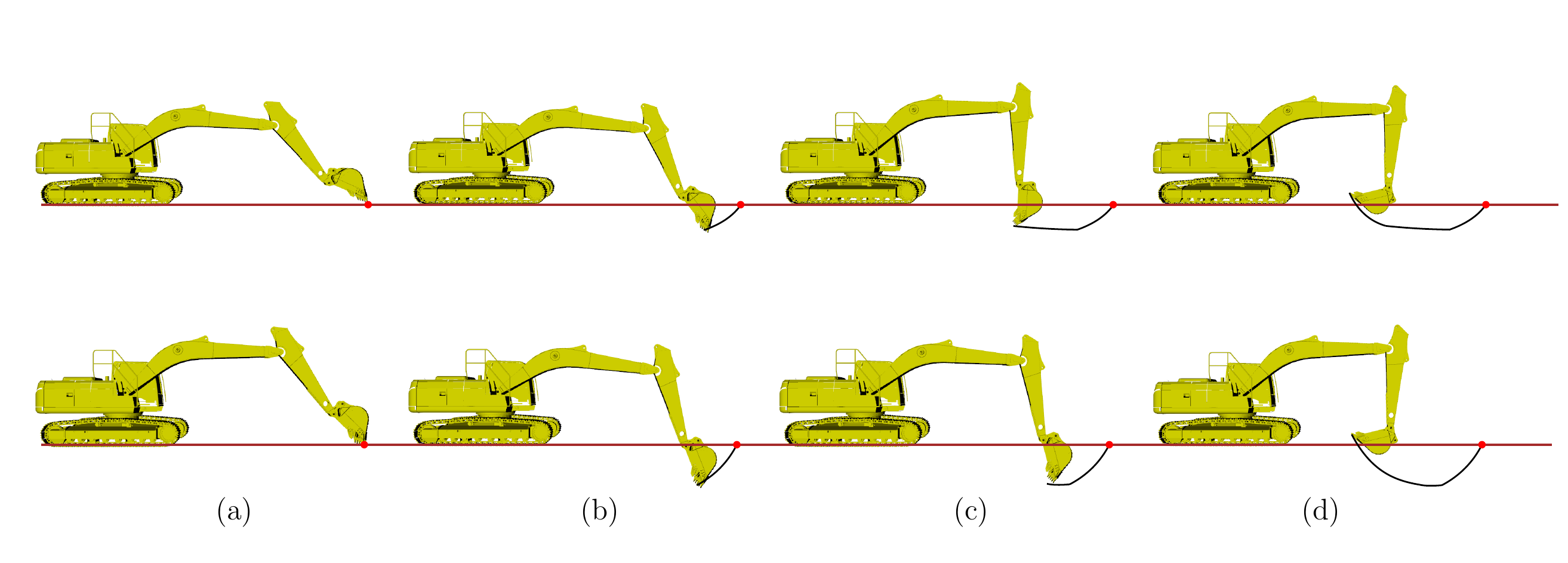}
    \caption{Two different bucket trajectories for excavation. (a) The bucket approaches the terrain surface. (b) The bucket penetrates into the soil. (c) The bucket drags horizontally to make the soil fail. (d) The bucket rotates and lift soils out from the terrain.}
    \label{fig:excvt_task}
\end{figure}

To fully fill the bucket with soils, one of the commonest excavation tasks, a trajectory of the bucket is softly constrained to follow a similar pattern of motions. Singh ~\cite{sing1995synthesis} concluded that the pattern is comprised of four phases: penetration, dragging, rotation and lifting. However, unlike other autonomous manipulation tasks such as welding and grinding in which end-effector paths are restricted to exactly desired ones, there exist many kinematically feasible bucket trajectory candidates satisfying the filling task for a given terrain, as shown in Figure~\ref{fig:excvt_task}. To select the dynamically optimal trajectory from as many trajectory candidates as possible, we need a proper and flexible way to formulate the excavation task instead of just providing an end-effector trajectory to track. Furthermore, since interaction forces are influenced by soil properties such as density and cohesion, trajectories should be adaptive to soil conditions for dynamics feasibility. In this paper, we focus on planning dynamically optimal trajectory for excavator considering the soil-tool interaction.

\subsection{Related Work}
 A motion planner for excavation needs to simultaneously consider the end-effector kinematics constraints and dynamics performances such as torque demands. There have been diverse research areas for control and planning of a single excavation operation ~\cite{singh1992task}. Jud~\emph{et al.}~\cite{jud2017planning} proposed an optimization-based force control method, which is adaptive to different soil conditions and protects the excavator from exceeding force/torque limits. The method optimizes for the next control command, but does not optimize over the entire excavation cycle to generate an optimal trajectory. Singh~\cite{sing1995synthesis} formulates the excavation motion planning problem as an optimization over a simplified variables~\emph{i.e.}, the bucket's approaching angle, height and digging distance, by determining other variables aforementioned with some deliberate metrics. However, since the formulation is overly simplified, it loses a lot of space for optimization. In \cite{kim2013dynamically}, time-efficient and minimum torque motions are generated with the classical parametric optimization method based on B-splines. Since they regard the digging phase as a point to point process without considering kinematics constraints, the computed trajectory might violate some excavation task specific constraints. In the motion planning literature, tree sampling-based planners with forward proportion have been successfully developed for planning dynamically feasible motion \cite{karaman11,lavalle01,bekris16}. While some of these approaches can achieve asymptotic optimality in theory, they tend to have slow convergence to high quality trajectories in practice.


\subsection{Main Contribution}

This paper proposes a minimal torque and time variable trajectory optimization method for a single excavation task. To reduce the number of optimization variables and improve the overall efficiency, we propose to only use few keypoints as variables and compute torque of linearly interpolated waypoints with small fixed time step between them. To enable autonomous time allocation, we also put the time intervals between keypoints under optimization. In this way, it is possible that the optimized trajectory not only costs less torque but also less time compared to the initial trajectory.

Our formulation considers geometric, kinematic and dynamics constraints for generating feasible excavation motion. We define swept volume geometric constraint to estimate the amount of soil to be excavated. The heading and movement directions of the bucket are further constrained over the entire trajectory. The common constraints of a robotic maniputlator,~\emph{e.g.} the joint angles, velocity and torque limits are also imposed. To efficiently calculate torque, we use the recursive inverse dynamics (ID) algorithm based on Lie group. To effectively take into account the interaction between the soil and the excavator bucket, we introduce a simple soil-tool interaction force model, which considers the geometric shape of the bucket and the most fundamental properties of the soil. Finally, we use a kinematic-based method for quick searching of initial trajectory for the optimization using a reachability map based representation.

As compared to previous work excavation motion planning and control   \cite{jud2017planning,kim2013dynamically,sing1995synthesis}, our optimization-based trajectory generation method has the following advantages:
\begin{enumerate}
    \item Our formulation explicitly takes into account complete excavation specific constraints in a straightforward way while most previous methods only consider a subset of these constraints. Using our method, physically feasible excavation trajectories are generated. 
    \item Unlike the parametric-based trajectory representation, which tends to be overly constrained for the underlying solution space, the keypoint-based optimization formulation is less conservative and can yield more optimal solution by exploring large state space.
    \item By using time variable keypoints, we can reduce the total number of optimization variables and thus improve the overall performance, which is achieving minimum torque while spending less time. The experiment shows that the optimization problem can quickly converge with few iterations and dynamically feasible trajectory can be computed within $150$ seconds on average.  
\end{enumerate}

We highlight the experimental result by comparing the trajectories generated under different swept volume, torque limit and other constraints, and demonstrate our method can generate dynamically feasible trajectories. 
We further validate the generated trajectories using AGX \footnote{https://www.algoryx.se/products/agx-dynamics/} - a high fidelity dynamic simulator with realistic terrain modeling. The experimental results further show that the trajectories can be successfully tracked with simple trajectory controller. Despite the soil-tool modeling difference between our method and the simulator, the torques along the entire excavation cycle are consistent between each other.  

The rest of the paper is organized as follows. Section 2 presents the excavation trajectory optimization problem and constraint formulation. Section 3 presents the bucket-soil interaction model. Section 4 presents experiments of our optimization approach. Conclusions and future work are in Section 5.




\section{Excavation Trajectory Optimization}



In this section, we present our time variable optimization-based excavation trajectory generation method. We will first introduce the classical optimal control problem from which the dynamical trajectory optimization formulation is derived. We then illustrate excavation task specific constraints which guarantee the trajectory is successful to excavate soils while leaving a plentiful room for optimization.

\subsection{Trajectory Optimization Statement}
The classical optimal control problem is to minimize some cost functionals over trajectory variables which could be formulated as:
\begin{align}
    &\minimize_{\tau(t)}\ \Phi(q(t), \dot{q}(t), t_f) + \int_0^{t_f}L(q(t), \dot{q}(t), \tau(t))dt \label{eq: opt_ctrl_cost} \\
    &\subjectto\ q(0) = q_0,\ \dot{q}(0) = 0,\ q(t_f) = q_{t_f}, \dot{q}(t_f) = 0 \label{eq: opt_ctrl_cstr}
\end{align}
where $q$ and $\dot{q}$ represent trajectory position and velocity in the joint space while $\tau$ is the necessary torque to execute the trajectory. The objective combines a main cost function $L$ and a penalty term $\Phi$ to force the final condition at $t_f$ to be satisfied. In general, the trajectory is constrained by the boundary conditions as described in Equation~\ref{eq: opt_ctrl_cstr}. The final time $t_f$ could be either fixed or flexible~\cite{lee2005newton}.

Torque and trajectory variables must obey the equation of motion:
\begin{equation}
    \tau = M(q)\ddot{q} + C(q, \dot{q})\dot{q} + V(q) - J_s^T(q)\mathcal{R}
    \label{eq: eom}
\end{equation}
where $M$ denotes the mass inertia of the manipulator, $C$ is the Coriolis matrix and $V$ contains gravitational forces. $\mathcal{R}$ is the external force usually applied at the end-effector of the manipulator.

For numerical computation, the trajectory is discretized into a sequence of joint positions $\mathbf{q} = \{q_0, q_1, \dots, q_N\}$ with $N+1$ timesteps. As a result, the minimal torque trajectory could be approximately expressed as
\begin{align}
    \minimize_{\mathbf{q}}\ \sum_{i = 0}^{N}||\tau_i||^2 \quad \subjectto\ \mathbf{g}(\mathbf{q}) \leq \mathbf{0}, \mathbf{h}(\mathbf{q}) = \mathbf{0}
\end{align}
where $\mathbf{g}$ and $\mathbf{h}$ represent a tuple of inequality and equality constraints for a specified task. Since many tasks in reality are actually not a point-to-point problem, we relax the boundary constraints commonly required in the optimal control formulation.
With a given small time step $\Delta t$ between two adjacent discrete points, the joint velocity and acceleration are obtained with
\begin{align}
    \dot{q}_i = \frac{q_{i+1} - q_i}{\Delta t} \quad\text{and}\quad
    \ddot{q}_i = \frac{\dot{q}_{i+1} - \dot{q}_i}{\Delta t}.
\end{align}
$\Delta t$ should be very small so that $\dot{q}$ and $\ddot{q}$ are calculated accurately enough to obtain the correct torque value. In other words, the number of variable waypoints $N+1$ should be large, which in turn increases the complexity of the optimization problem.

To deal with this issue, we propose a concept of keypoints of the trajectory, based on the observation that, for tasks like excavation (as will be shown in the following section), constraints on the entire trajectory could be reduced to constraints on some keypoints. We only use keypoints as variables of the optimization problem and compute torque of linearly interpolated waypoints with fixed time step $\Delta t$ between keypoints. The problem is then reformulated as
\begin{align}
    \minimize_{\mathbf{k}}\ \sum_{i = 0}^{M} \sum_{j = 0}^{Q}||\tau_{i, j}||^2 \quad \subjectto\ \mathbf{g}(\mathbf{k}) \leq \mathbf{0}, \mathbf{h}(\mathbf{k}) = \mathbf{0}
\end{align}
where $\mathbf{k} = \{k_0, k_1, \dots, k_M\}$  denotes selected keypoints with $M$ size and $Q$ is the number of interpolated waypoints.


To enable flexible final time $t_f$ as in the aforementioned optimal control problem, we add the time intervals between keypoints $\mathbf{T} = \{T_0, T_1, \dots, T_{M-1}\}$ as variables in the optimization problem. Subsequently, $Q_i = \frac{T_i}{\Delta t}$ is no longer fixed. Finally, we adopt the problem formulation:
\begin{align}
    \minimize_{\mathbf{k}, \mathbf{T}}\ \sum_{i = 0}^{M} \sum_{j = 0}^{Q_i}||\tau_{i, j}||^2 \quad \subjectto\ \mathbf{g}(\mathbf{k},\mathbf{T}) \leq \mathbf{0}, \mathbf{h}(\mathbf{k}, \mathbf{T}) = \mathbf{0}
\end{align}


\subsection{Excavator Model}
\begin{figure}[htb]
    \centering
    \includegraphics[width=0.7\linewidth]{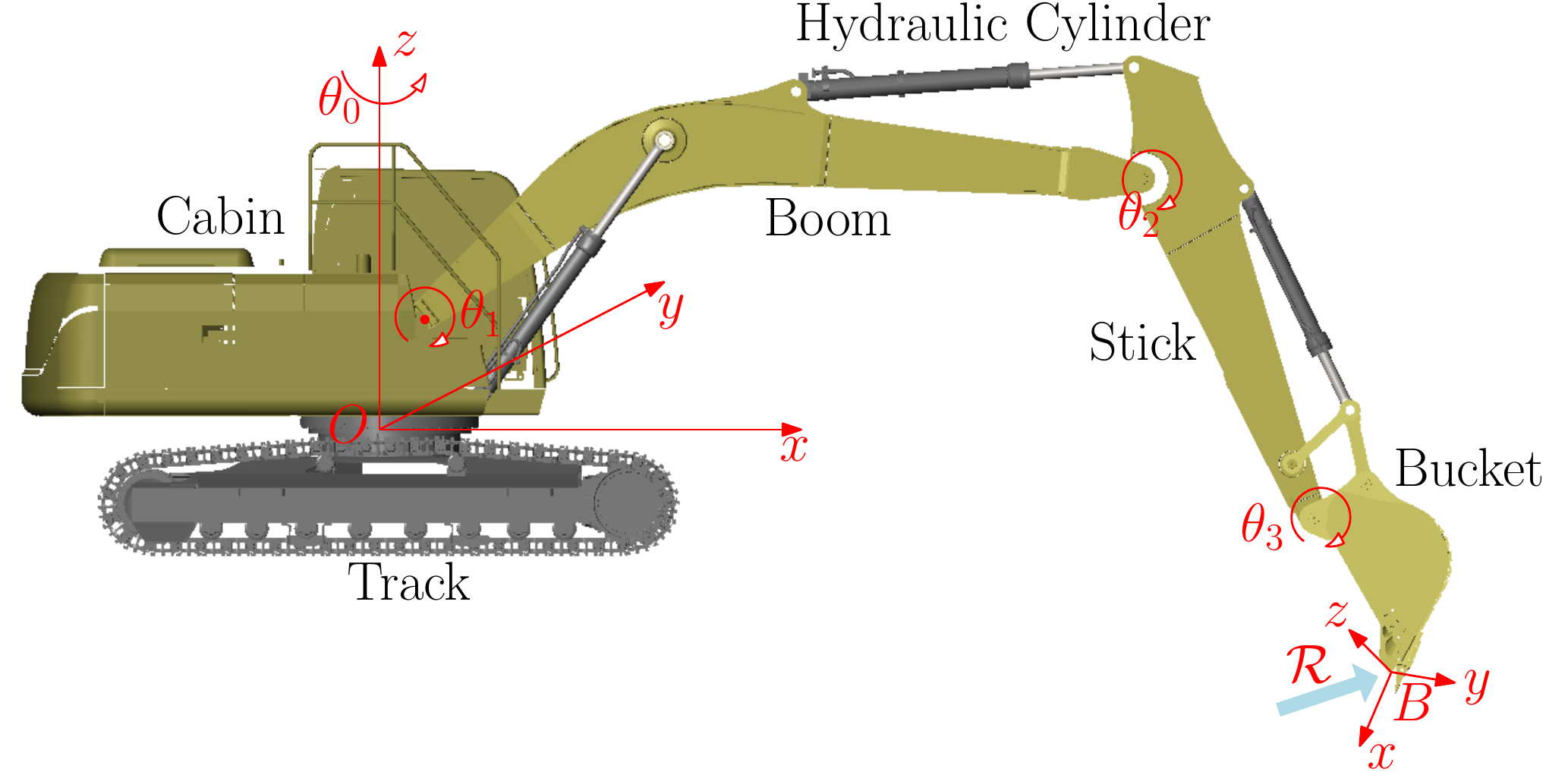}
    \caption{Excavator Model}
    \label{fig:excvt_model}
\end{figure}

As Figure~\ref{fig:excvt_model} describes, an excavator is a hybrid serial-parallel mechanism. It is composed of serially connected links: cabin, boom, stick and bucket. Hydraulic cylinders are assembled on these links which result in several closed-loop structures. The prismatic hydraulic cylinder forces could be transformed into torque of $4$ revolute joints according to the excavator's geometric dimensions~\cite{jud2017planning, kim2013dynamically}. Thus, for the excavation problem, the commonly generalized joint position variable $q = [\theta_0, \theta_1, \theta_2, \theta_3]^T$ is utilized.

The body fixed coordinate frame $B-xyz$ located at the tip of bucket represents bucket pose $g_B = (R_B, p_B)$ relative to the track base fixed coordinate frame $O-xyz$, where $R_B$ denotes orientation and $p_B$ is the position of the tip. Given a joint position, the bucket pose is calculated with the forward kinematics function: $g_B = \FK(q)$.

During an excavation, the resistance forces that surrounding soils exerting on the bucket surface could be integrated into one resultant force $\mathcal{R}$ applied at $B-xyz$. For each waypoint on the trajectory, we calculate the necessary torque to achieve the target acceleration while overcoming $\mathcal{R}$ with the parallel recursive inverse dynamics algorithm~\cite{park1995lie,yang2017parallel}.

\subsection{Excavation Constraints}
\begin{figure}[htb]
    \centering
    \includegraphics[width=.5\linewidth]{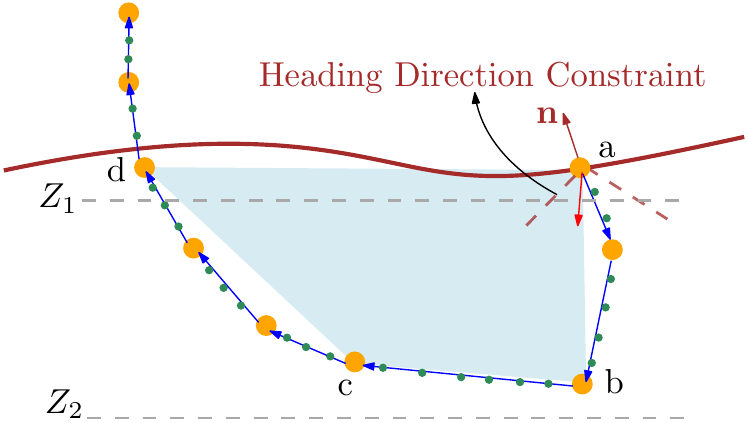}
    \caption{Excavation constraints. Large orange circles denote bucket tip keypoints. Between each pair of adjacent keypoints, waypoints, denoted by tiny green circles, are linearly interpolated. Translation and heading directions are drawn with blue and red directions. The brown curve represents the terrain surface. $\mathbf{n}$ is the normal vector of the surface at the keypoint a. The heading direction is restricted between a range around $\mathbf{n}$. The blue shadow area denotes the side view of the excavated swept volume. Depths of keypoints b and c are forced to be within the range $[Z_1, Z_2]$ to guarantee that they are beneath the soil surface.}
    \label{fig:excvt_cstr}
\end{figure}

In this section, we will illustrate excavation constraints and how to select keypoints to reduce the computation complexity. Constraints play two roles in an optimization problem for a given task:
\begin{enumerate}
    \item They specify the requirement of the task, while the optimization objective function represents the criteria of how well the task is performed.
    \item They describe other limitations and rules for the robot to obey.
\end{enumerate}
For the excavation task, an excavator needs to fill the bucket with a desired mass of soils, while not violating other constraints such as torque limits the machine could exert and the range of bucket path due to target terrain shape. Fig.~\ref{fig:excvt_cstr} illustrates keypoints selection and some important constraints as an example for the excavation task. The bucekt penetrates into the soil from a to b; drags from b to c; rotates from c to d. Compared to other movements, more constraints are imposed to the rotation. For example, the heading directions and translation directions should coordinate well to swept soils in the bucket. Thus, more keypoints are used for the rotation phase. Please refer to notes of the figure for detailed explanation.

\subsubsection{Swept Volume Constraint}
Although the bucket filling is a complex dynamics process depending on both soil conditions and the bucket trajectory~\cite{coetzee2007discrete}, the excavated soils could be estimated with the swept volume --- the soil volume above the bucket path~\cite{jud2017planning, kim2013dynamically}. As shown in Figure~\ref{fig:excvt_cstr}, the swept volume could be approximated with four points: the entry point of soil $k_0$ (a), end point of the penetration phase $k_1$ (b); the end point of the dragging phase $k_2$ (c), and the exit point of the soil $k_4$ (d). we enforce the swept volume to be within a range determining by the bucket's real capacity:
\begin{equation}
    V_{\min} \leq V \leq V_{\max},\ \quad V = S_{abcd} \times W
\end{equation}
where $W$ is the bucket width perpendicular to the excavation plane and $S_{abcd}$ is the area of the quadrilateral $abcd$.

\subsubsection{Direction Constraints}
\begin{figure}[htb]
    \centering
    \begin{subfigure}{0.49\textwidth}
      \centering
      \includegraphics[width=.3\linewidth]{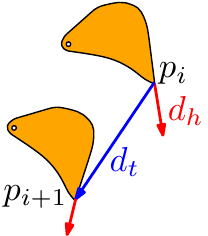}
      \caption{Directions}
      \label{fig:directions}
    \end{subfigure}
    \begin{subfigure}{0.49\textwidth}
      \centering
      \includegraphics[width=.5\linewidth]{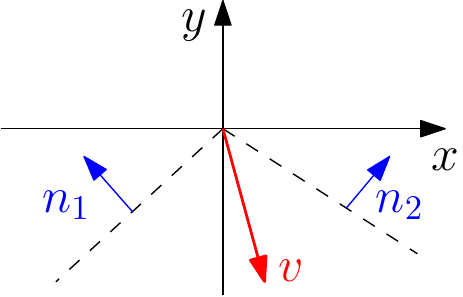}
      \caption{Direction Constraints}
      \label{fig:dir_cstr}
    \end{subfigure}
    \caption{(a) illustrates the heading direction $d_h$ and bucket tip translation vector $d_t$, which are two important factors for the success of an excavation task. 
    (b) illustrates direction constraints represented using hyperplanes. For example, the direction of $v$ is constrained within the intersection of half-planes defined by vector $n_1$ and $n_2$, i.e. $v\cdot n_1 <0$ and $v\cdot n_2<0$.}
    \label{fig:direction_and_cstr}
\end{figure}

Whether a bucket trajectory is successful for an excavation task highly depends on two directions. As Figure~\ref{fig:directions} depicts, one is the bucket heading direction $d_h$ which refers to the direction along the bucket tooth; the other is the bucket tip translation vector $d_t$ referring to the bucket position movement between two adjacent time steps.
In general, as illustrated in Figure~\ref{fig:dir_cstr}, we use hyperplanes to restrict directions. The vector $v$ locating in the half-plane which is in the opposite direction of the normal vector $n$ could be expressed as $v \cdot n < 0$.

\begin{figure}
    \centering
    \includegraphics[width=.4\linewidth]{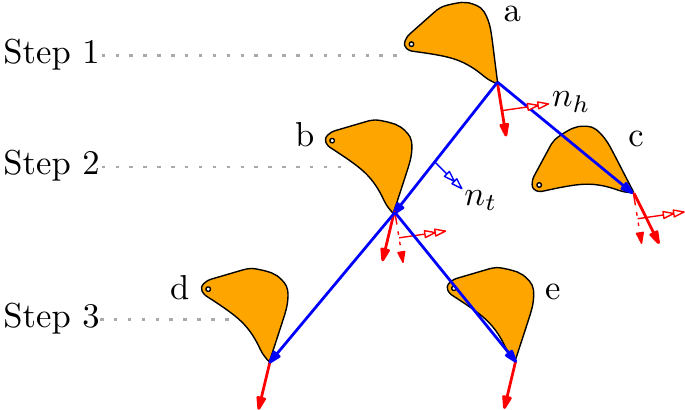}
    \caption{Monotonic change of the directions. The bucket moves from step $1$ to step $3$. Red arrows denote heading directions and blue arrows are translation directions. Double arrows represent the normal vectors of hyperplanes for corresponding directions. At step $2$, pose c is not valid since its heading direction locates at the forbidden halfplane. At step $3$, pose e violates the constraint of monotonic change of translation directions. As a result, the path $a \rightarrow b \rightarrow d$ is a feasible path.}
    \label{fig:dir_mono_change}
\end{figure}

The heading directions at some time steps are restricted within some ranges. For example, $d_h$ at the approach point is supposed to align with the normal vector of the terrain surface; $d_h$ during bucket lifting should keep approximately upright. With a certain heading direction $d_h^B$ with respect to the frame $B$, one could obtain its coordinate in the inertial frame $O$ with rotation $R$: $d_h = R(q)d_h^B$. Thus the constraints on heading direction and translation direction could be formulated as
\begin{equation}
    R(q) d_h^B \cdot n < 0,\ \quad d_t \cdot n = (p_{i+1}(q) - p_i(q)) \cdot n < 0
\end{equation}
For a successful excavation process, both heading directions and translation directions should not swing under the soil surface. Otherwise, the bucket will push the soil. In other words, these directions along the entire trajectory should monotonically change. We restrict directions by the previous one using the aforementioned hyperplane method, as illustrated in Figure~\ref{fig:dir_mono_change}.
The constraints of monotonic change of directions are formulated as
\begin{align}
    &d_{h\_i+1} \cdot R_{90} d_{h\_i} < 0 \text{ for } i = 0, 1, \dots, M \\
    &d_{t\_i+1} \cdot R_{90} d_{t\_i} < 0 \text{ for } i = 0, 1, \dots, M-1
\end{align}
where $R_{90}$ is rotation around the axis perpendicular to the excavation plane.
\subsubsection{Velocity and Torque Constraints}
In addition to the task specific constraints presented above, there are some constraints induced by the mechanical properties. We list two main constraints common in excavators: for all waypoints, motion should satisfy joint velocity constraints due to oil flow limits and joint torque constraints:
\begin{equation}
    \norm{\frac{q_{i+1} - q_i}{\Delta t}}^2  < v_{\max}^2 \quad\text{and}\quad \norm{\tau_i}^2 < \tau_{\max}^2.
\end{equation}

\section{Bucket-Soil Interaction Force}
This section considers the soil and the bucket tool interaction,~\emph{i.e.} the relationship between soil resistance force and the tool action. The interaction between the soil and tool is rather complicated as it is affected by the soil material property, the terrain deformation and other unknown parameters. Researches have put a lot of effort into developing mechanics quantitatively describing the reaction of soil to tools~\cite{gill1967soil,mckyes1985soil}. Analytical models such as the fundamental earthmoving equation~\cite{reece1964paper} has been developed with soil failure mechanics specifically for tillage or traction movements. However, they are not suitable for other types of movements such as penetration and pushing that might happen during the optimization process. From the visco-plastic nature and mass deformation of the soil, another genre of researchers utilize the fluid dynamics to model the tool soil interaction process~\cite{andreotti2013granular,karmakar2006dynamic, sauret2014bulldozing}. Inspired by them, we propose a simplified soil-tool force model based on the fluid dynamics, which maps from a joint state and action into the resistance force exerted by the surrounding soils $f(q, \dot{q}, \ddot{q}) \mapsto \mathcal{R}$.
\begin{figure}
    \centering
    \includegraphics[width=.8\linewidth]{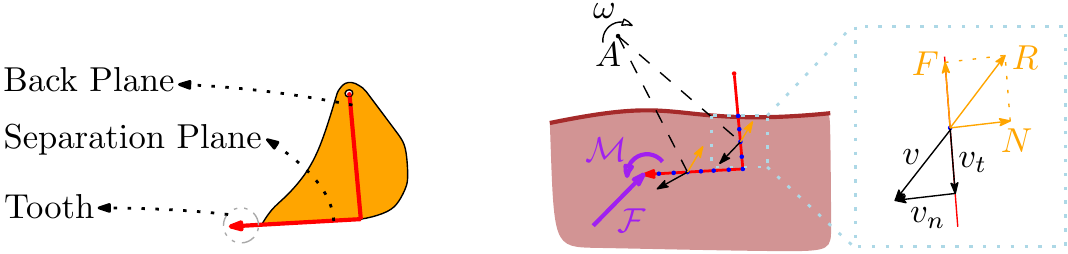}
    \caption{Bucket simplification and soil-tool interaction force model. The brown curve and shadow area beneath it represent the terrain soil. Any planar rigid body motion could be regarded as a rotation around some point $A$. Based on the rotation motion, velocity of each point is calculated and denoted with a black arrow $v$, which is decomposed into $v_n$ and $v_t$ shown in the enlarged right part of figure. $\mathcal{F}$ and $\mathcal{M}$ denote the resultant force and moment at the bucket tip frame.}
    \label{fig:force_model}
\end{figure}

As shown in Fig.~\ref{fig:force_model}, our model simplifies the bucket with two line segments, named the back plane and the separation plane~\cite{park2002development}. We assume that the soil pressure exerting on the parts of a static bucket surrounded by soils could be calculated with hydrostatic pressure formula:
\begin{equation}
    P = \rho g h
\end{equation}
where $\rho$ represents the mass density of the soil which is one of the most fundamental soil properties, $g$ denotes the gravitational constant, and $h$ represents the depth of the part from the soil surface. This formula indicates that resistance force from the soil will grow as the bucket penetrates more deeply, which is verified by experiments in~\cite{kang2018archimedes}. To make any movement, the bucket needs to overcome friction forces whose directions are opposite to the its velocity. As in general cases, the friction magnitude is given by:
\begin{equation}
    F = \mu N
\end{equation}
where $\mu$ denotes the friction coefficient and $N$ is the normal force applied at the bucket face. In this paper, a resistance force at a small area is comprised with the normal force and friction force.

We first compute resistance forces ($R_i = N_i + F_i$) for discretized small areas of the submerged bucket part and integrate them into the resultant force and moment $\mathcal{R} = (\mathcal{F}, \mathcal{M}$) at the bucket tip frame $B-xyz$.
We model both $N_i$ and $F_i$ as functions of the velocity of the $i$th small area. As the Figure~\ref{fig:force_model} illustrates, velocity of each small area $v = v_n + v_t$, where $v_n$ is normal to the bucket face while $v_t$ is along with this face. $N$ and $F$ are opposite to $v_n$ and $v_t$ respectively. The force magnitude is controlled by several coefficients that implicitly stand for the soil conditions. In particular, $N$ for each area is given by:
\begin{equation}
    N = -\frac{(K_{p}P + K_v \norm{v_n})S}{\norm{v_n}}v_n
\end{equation}
where $S$ is the area of the element segment. $K_p$ scales the height related pressure while $K_v$ adjusts the influence of velocity magnitude.
$F_s$ is calculated with
\begin{equation}
    F_{s} = -\frac{K_sN}{\norm{v_t}}v_t
\end{equation}
where $K_f$ is the friction coefficient depending on the soil cohesion and density~\cite{mckyes1985soil}.

We obtain the velocity of each point fixed to the bucket with its body velocity $V_B = (\omega_B, v_B)$ which is calculated with the differential forward kinematics
\begin{align}
    V_B = J(q)\dot{q} \quad\text{and}\quad v_i = r_i \times \omega_B + v_B
\end{align}
where $\omega_B$ is the angular velocity of the frame $B$, $v_B$ is its linear velocity, $J$ is the kinematics Jacobian matrix, $r_i$ is the point position relative to the frame $B$. Finally, the resultant force $\mathcal{F}$ and moment $\mathcal{M}$ is computed by
\begin{align}
    \mathcal{F} = \sum_{i} R_i \quad\text{and}\quad \mathcal{M} = \sum_{i} r_i \times R_i
\end{align}
\section{Experiments}
In this section, we present the implementation and experiment results of our optimization-based trajectory generation. Our method uses the open source TrajOpt SQP solver~\cite{schulman2014motion} to solve the optimization problem. We efficiently calculate torque with the recursive inverse dynamics (ID) algorithm based on Lie group \cite{yang2017parallel}. We use a kinematic-based method for quick searching of initial trajectory for the optimization using a reachability map based representation \cite{yang2019compact}. In our experiments, we show that our approach can generate trajectories satisfying various constraints such as swept volume constraint, and adaptive to different soil property. We further demosntrate that, in some cases, with variable time intervals between keypoints, our optimization method converges faster and also reduces the cost. 
In order to further validate our approach, we use a commercial dynamic simulator AGX to simulate and validate the generated trajectories. Important parameters and mechanical constraints of the excavator platform are listed in Table~\ref{tab:excvt_specs}.

\begin{table}[ht]
\centering
\begin{tabular}{|a|a|a|a|p|p|}
\hline
       & Mass (\SI{}{\kilogram}) & Torque Limit (\SI{}{\newton\meter}) & Velocity Limit (\SI{}{\radian\per\second}) & \multicolumn{2}{p|}{Bucket Size (\SI{}{\meter})}  \\ \hline
Boom   & $1434.52$  & $950000$ & $0.785$ & Width & $1$  \\ \hline
Stick  & $656.189$ & $425000$ & $0.785$ & Length &  $1$ \\  \hline
Bucket & $809.59$ & $300000$ & $0.785$ & Depth &  $1$ \\ \hline
\end{tabular}
\caption{Excavator Specifications}
\label{tab:excvt_specs}
\end{table}


\subsubsection{Constraint Satisfaction}
This experiment demonstrates the basic function of our algorithm: given an initial trajectory that might violate some constraints, the motion planner could generate a feasible trajectory that meets the requirement of swept volume while minimizing the torque cost. In Experiment \Rmnum 1, the excavator is required to fill the bucket with soils of volume $\SI{0.8}{m^3} \leq V \leq \SI{1}{m^3}$. Note that since the swept volume is an approximate value and it tends to be smaller than the actual excavated volume, we set this constraint less than $\SI{1}{m^3}$.

\begin{figure}
    \centering
    \begin{subfigure}{0.49\textwidth}
      \centering
      \includegraphics[width=\linewidth]{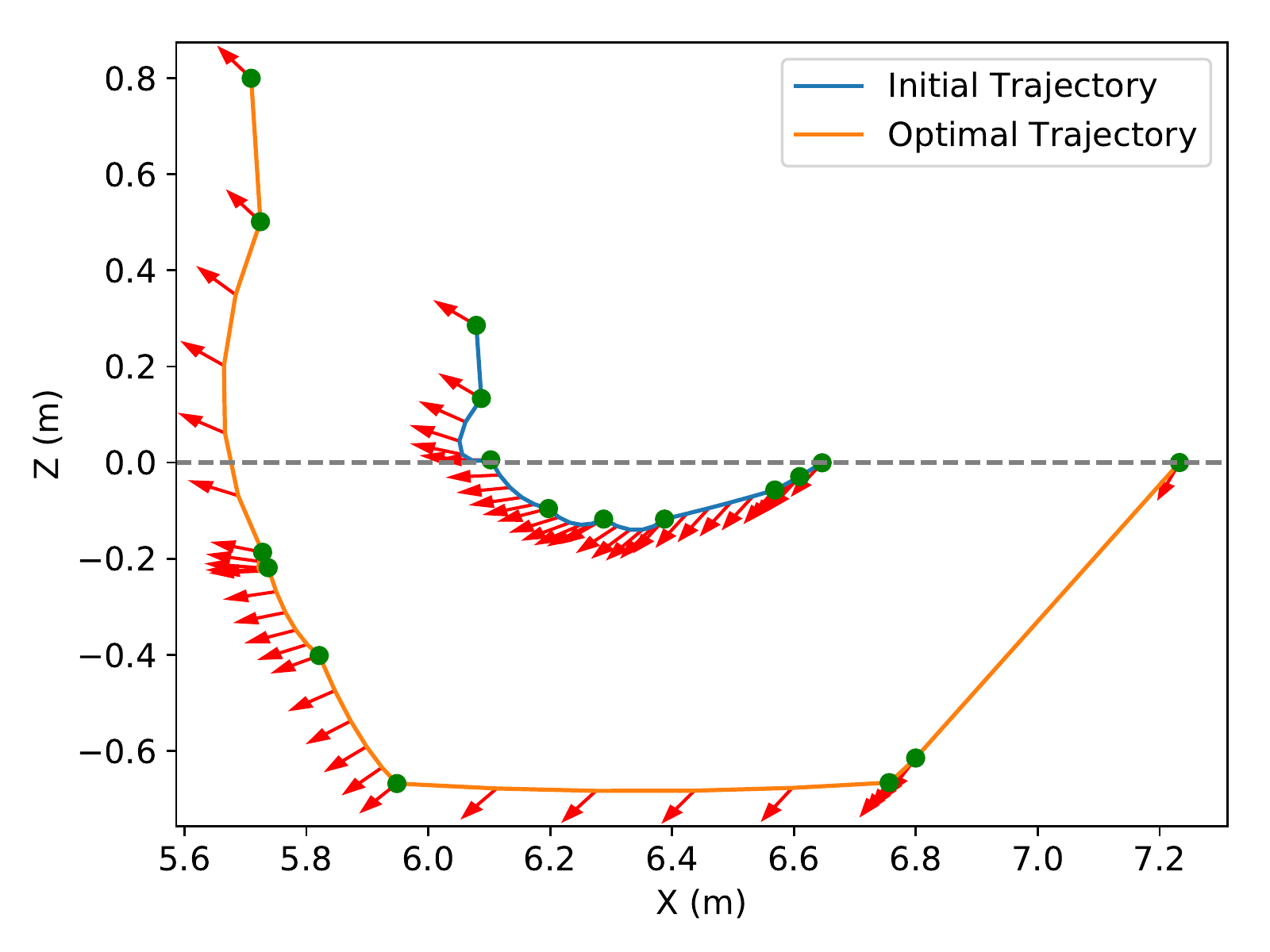}
      \caption{Bucket Trajectories}
      \label{fig:EXP_1_bkt_traj}
    \end{subfigure}
    \begin{subfigure}{0.49\textwidth}
      \centering
      \includegraphics[width=\linewidth]{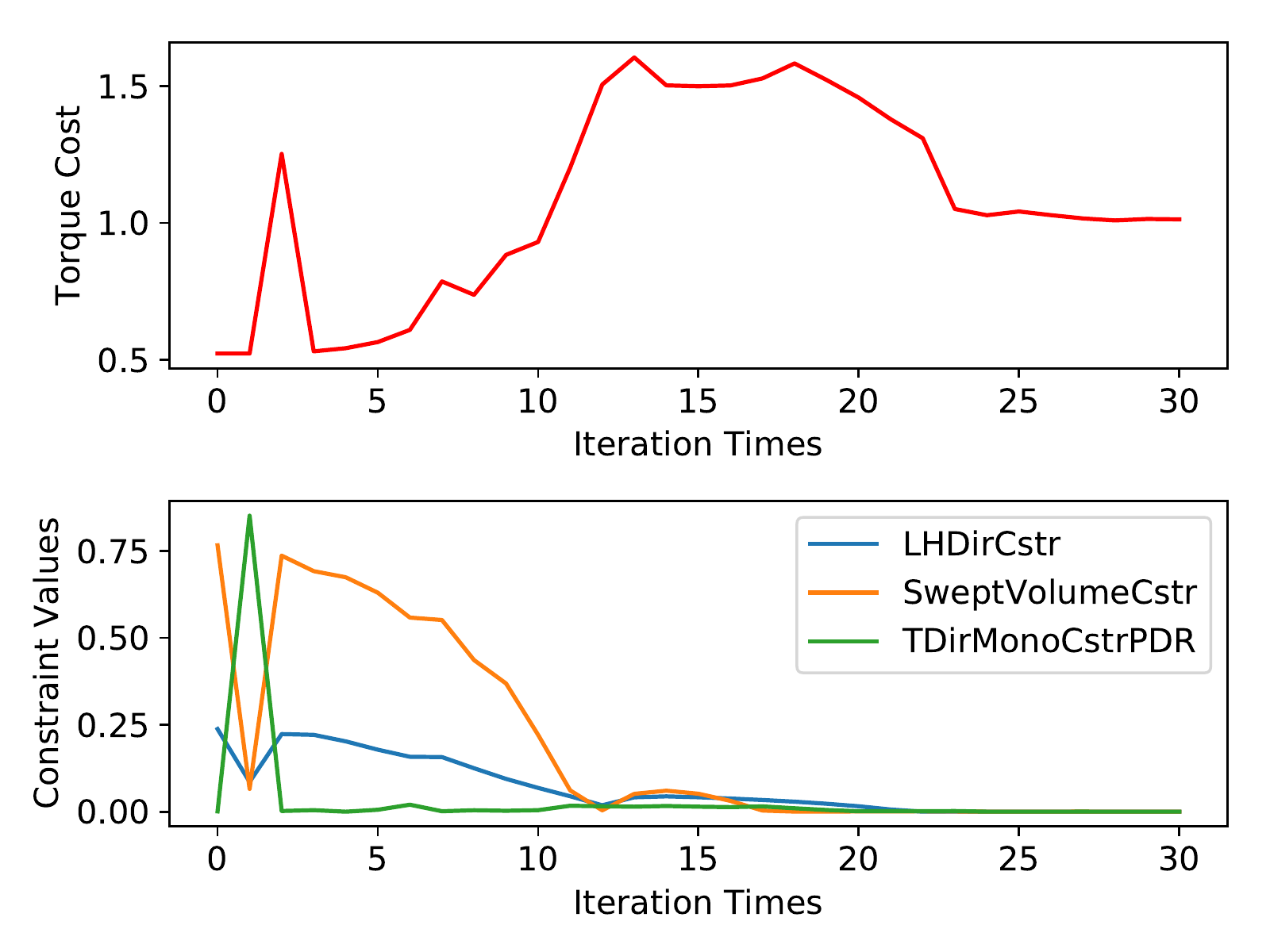}
      \caption{Torque Cost and Constraint Values}
      \label{fig:EXP_1_cost_cstr}
    \end{subfigure}
    \caption{Experiment \Rmnum 1. The sub-figure (a) shows the side view of the bucket trajectory which is parallel to the excavation plane. Each dot point denotes a keypoint at which the heading direction is represented by the red arrow. The sub-figure (b) draws values of torque cost and some constraints during optimization. LHDirCstr is the heading direction constraint of lifting phase; SweptVolumeCstr is the constraint of swept volume for the whole trajectory; TDirMonoCstrPDR is the monotonic change constraint of the translation direction for penetration-dragging-rotation phases.}
    \label{fig:EXP_1}
\end{figure}

The swept volume of the initial trajectory, as shown in Figs.~\ref{fig:EXP_1}, is apparently less than $\SI{0.8}{\meter^3}$. Thus, the value of volume constraint is large when the optimization starts. Time intervals between all keypoints are $\SI{1}{\second}$ and we fix them during optimization in this experiment. It is expectable that the torque cost increases since the excavator is required to dig more soils. After all constraints are satisfied, the cost continues to decrease until convergence.

\begin{figure}[h]
    \centering
    \begin{subfigure}{.49\textwidth}
    \centering
    \includegraphics[width=\linewidth]{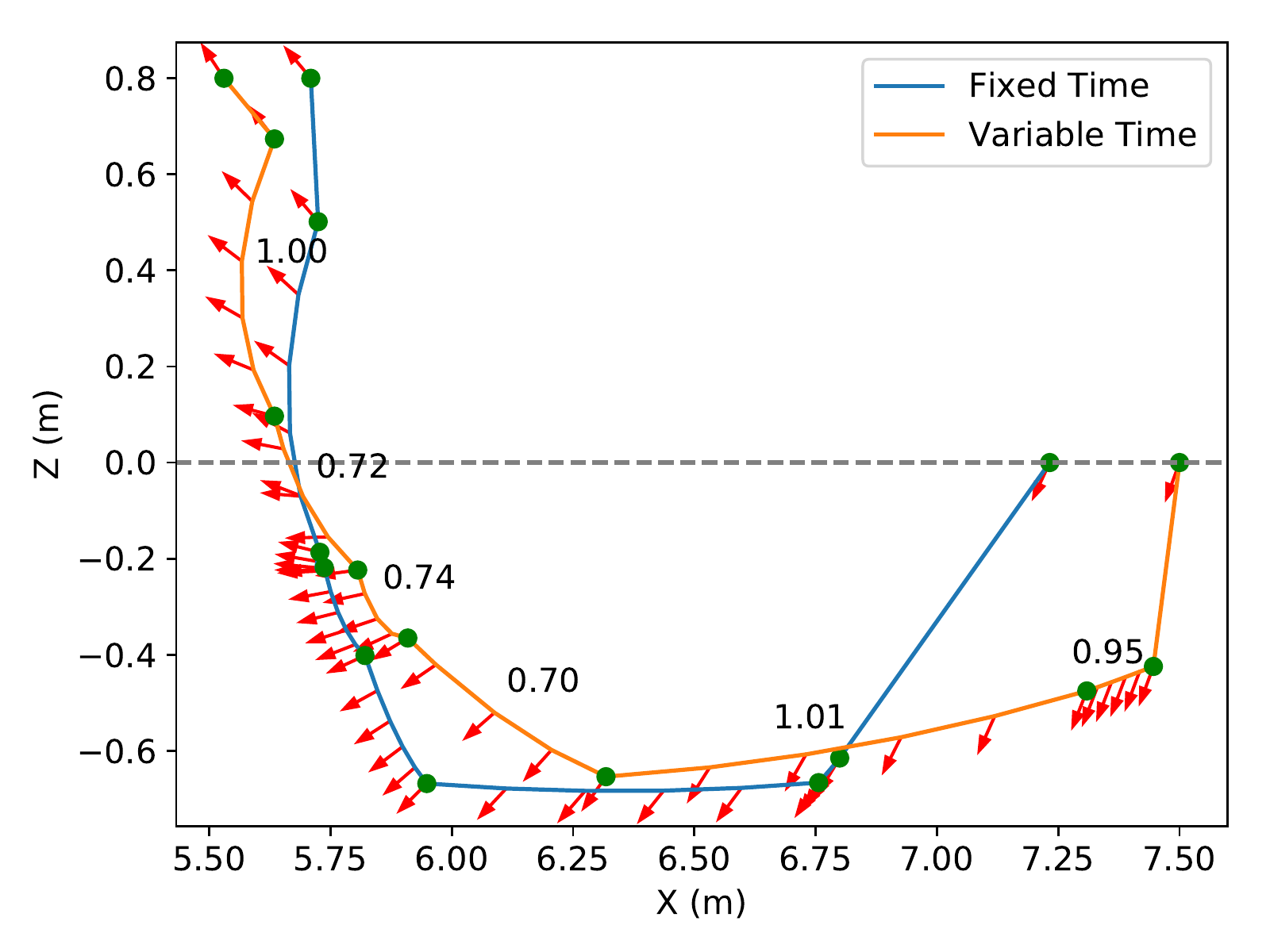}
    \caption{Bucket Trajectories}
    \end{subfigure}
    \begin{subfigure}{.49\textwidth}
    \centering
    \includegraphics[width=\linewidth]{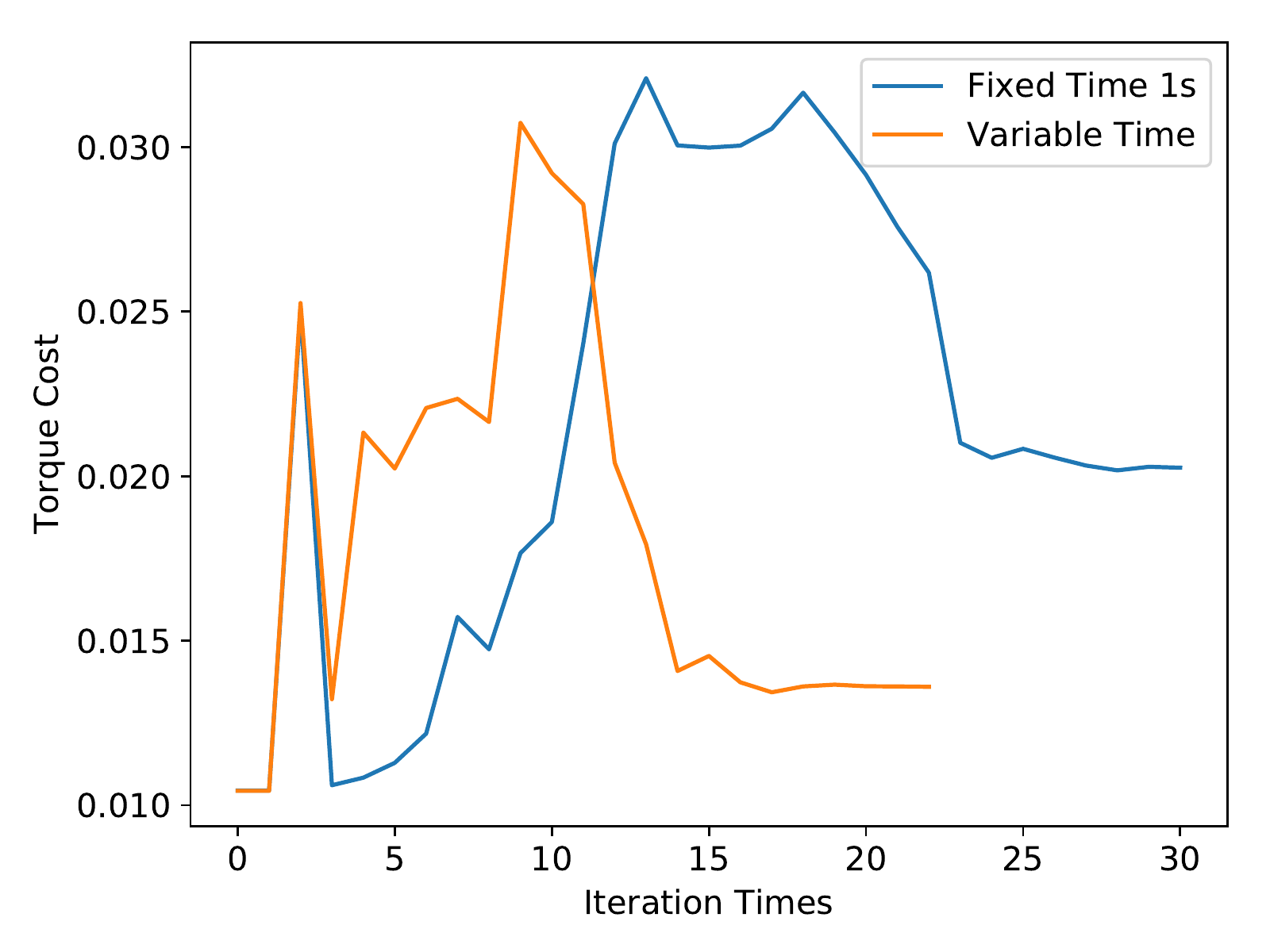}
    \caption{Torque Cost}
    \end{subfigure}
    \caption{Comparison of optimization results between fixed and variable time experiments. In sub-figure (a), time intervals between keypoints are annotated for the time variable trajectory.}
    \label{fig:new_EXP_2}
\end{figure}

\subsubsection{Variable Time}
We further conduct an experiment with variable time intervals between keypoints and find that time scaling could help the optimization converge faster and also reduce more cost. As annotated in Fig.~\ref{fig:new_EXP_2}, the time variable trajectory totally spends $\SI{5.5}{\second}$ while the time fixed trajectory spends $\SI{6.4}{\second}$. To find a minimal sum of torque over a trajectory is to find a balance between short time and small acceleration. Short time means few number of torques to be added, while small acceleration requires less torque for a single time step according to the equation of motion. This balance basically depends on several factors, such as the mass properties of the excavator and the resistance force exerted by soils. Therefore, allowing time intervals variable actually provides extra dimension to diminishing sum of torque.

\subsubsection{Different Soil}
To demonstrate this method could adapt to different soil conditions, we compare trajectories for two types of soils, whose properties are given in Table~\ref{tab:soil_properties} with the same initial trajectory. Since this experiment aims to emphasize the capability of adpation to different end-effecor force, we use an initial trajectory without violating constraints. That means, the trajectory adaptation is purely determined by the process of minimizing torque costs, which shows that the inverse dynamics objective function is well conditioned for optimization. As shown in Fig.~\ref{fig:new_EXP_3}, the trajectory for the hard soil is shallower than the one for the soft soil. This result matches the trajectory executed by the human operators~\cite{jud2017planning}.


\begin{table}[h]
\begin{minipage}[b]{0.45\linewidth}
\centering
\resizebox{\linewidth}{!}{
\begin{tabular}{|c|c|c|}
\hline
& Soft Soil & Hard Soil \\ \hline
$\rho\ (\SI{}{kg.m^{-3}})$ & $1000$ & $2500$ \\ \hline
$K_p$ & $1$ & $3$ \\ \hline
$K_v$ & $300$ & $1000$ \\ \hline
$K_s$ & $0.5$ & $0.8$ \\ \hline
\end{tabular}
}
\vspace{10pt}
\caption{Soil Properties. We simultaneously change density and coefficients which are influenced by other attributes to simulate different soil conditions.}
\label{tab:soil_properties}
\end{minipage}\hfill
\begin{minipage}[b]{0.54\linewidth}
\centering
\includegraphics[width=\linewidth]{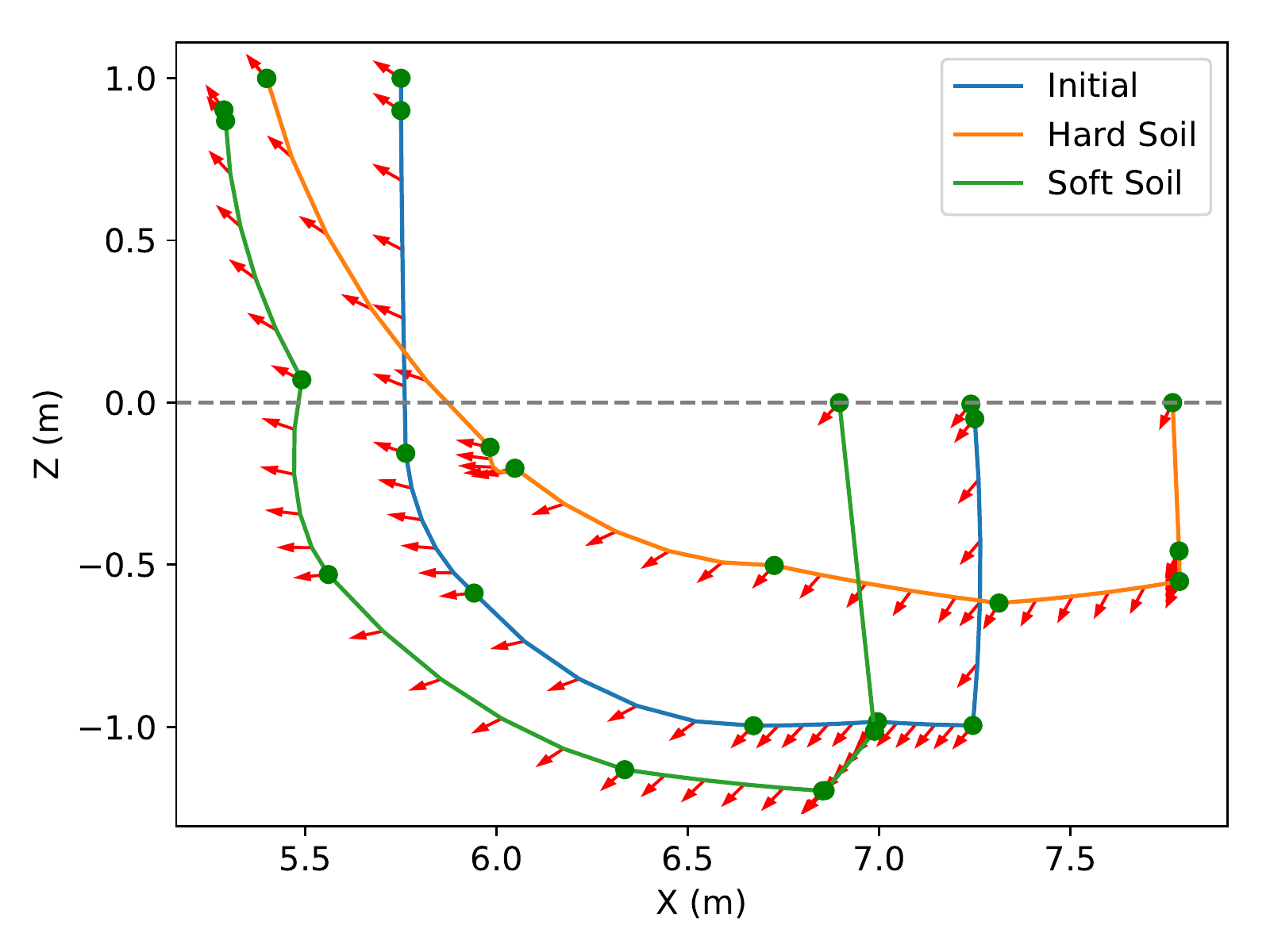}
\captionof{figure}{Bucket trajectory comparison}
\label{fig:new_EXP_3}
\end{minipage}
\end{table}

\subsubsection{Validation on Simulator}
\begin{figure}[h]
    \centering
    \begin{subfigure}{.49\textwidth}
    \centering
    \includegraphics[width=\linewidth]{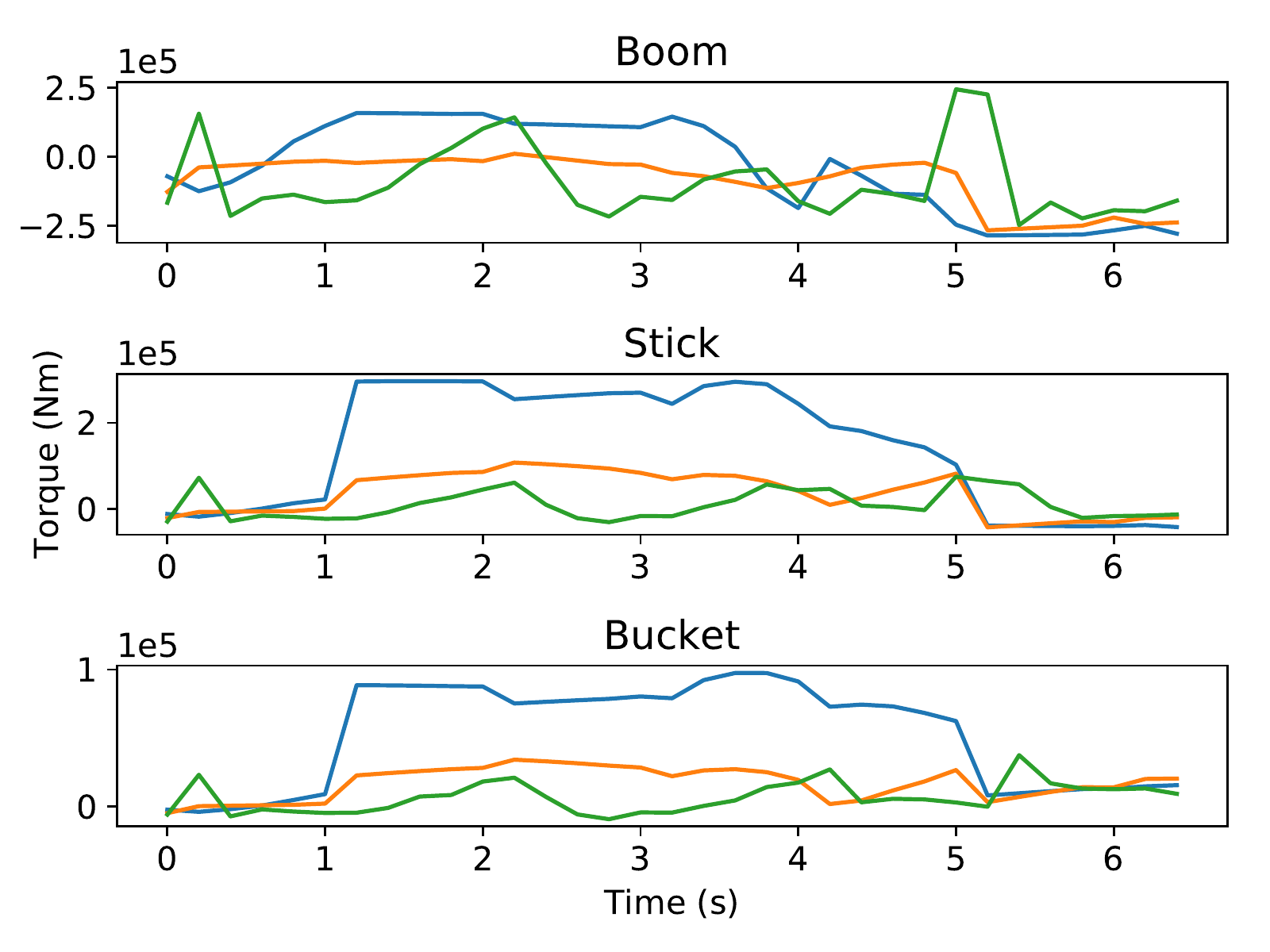}
    \caption{Torque}
    \end{subfigure}
    \begin{subfigure}{.49\textwidth}
    \centering
    \includegraphics[width=\linewidth]{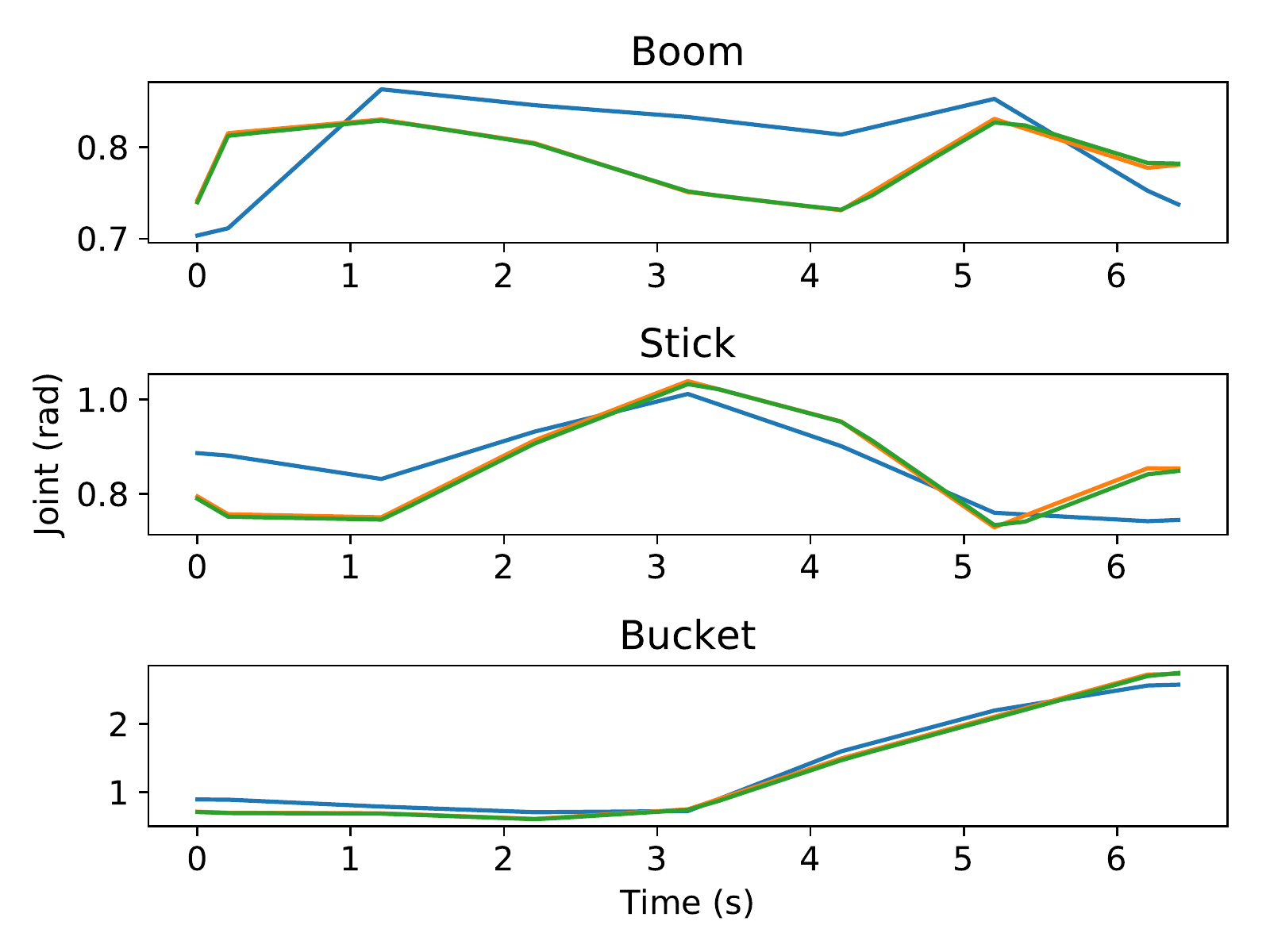}
    \caption{Joint Angle}
    \end{subfigure}
    \caption{Comparison of optimization results on dynamic simulator. Blue curves are for the initial trajectory; Orange curves are for the optimal trajectory; Green curves are for the optimal simulation trajectory.}
    \label{fig:torque_comp}
\end{figure}

\begin{figure}
    \centering
    \includegraphics[width=\linewidth]{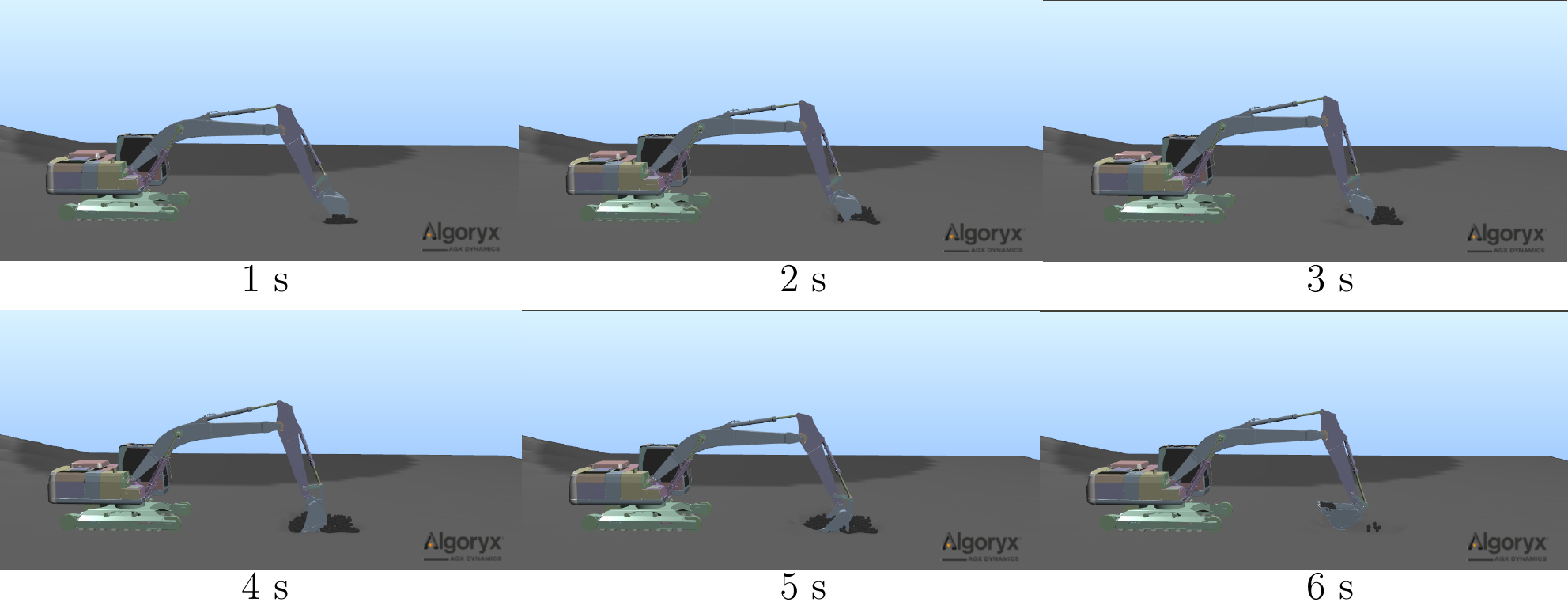}
    \caption{Sequence of excavation motion on dynamic simulator.}
    \label{fig:sim}
\end{figure}

We validate the generated trajectories using AGX - a high fidelity dynamic simulator with realistic terrain modeling. As shown in Fig.~\ref{fig:sim}, the trajectories can be successfully tracked with simple trajectory controller. Though there is difference of soil-tool model between our method and the simulator, the torques along the entire excavation cycle are consistent between each other as shown in Fig.~\ref{fig:torque_comp}.
 
\section{Conclusion}

This paper presents a minimal torque and time variable trajectory optimization method for a single excavation task. Our formulation considers geometric, kinematic and dynamics constraints for generating feasible excavation motion. Different from B-spline based parameteric optimization, our method directly optimizes the keypoints along trajectory and we propose a time variable mechanism where the time intervals between the keypoints are also under the optimization. We further introduce a soil-tool interaction force model to effectively take into account the interaction between the soil and the excavator bucket. We highlight the experimental result by comparing the trajectories generated under different swept volume and soil properties, and further validate the generated trajectories using a dynamic simulator. 

There are many avenues for future work. We are interested in integrating our excavator trajectory planner with task planner for high level excavation tasks. We are also interested in applying system identification to correlate and improve our soil-force model with the real-world excavation data on different soil environment. Finally, we are interested in testing our method on real excavators.

\bibliographystyle{plainnat}
\bibliography{ref}
\end{document}